\documentclass[10pt,twocolumn]{article}

\usepackage[
  letterpaper,
  left=0.75in,
  right=0.75in,
  top=1.0in,
  bottom=1.0in,
  columnsep=0.25in
]{geometry}

\usepackage[T1]{fontenc}
\usepackage{times}
\usepackage{microtype}
\usepackage{ragged2e}

\usepackage{amsmath,amssymb,amsthm,mathtools}
\usepackage{graphicx}
\usepackage{float}
\usepackage{subfigure}
\usepackage{booktabs}
\usepackage{hyperref}
\usepackage[capitalize,noabbrev]{cleveref}
\usepackage{placeins}
\usepackage{dblfloatfix}

\usepackage{titlesec}
\titleformat{\section}{\large\bfseries}{\thesection}{0.5em}{}
\titleformat{\subsection}{\normalsize\bfseries}{\thesubsection}{0.5em}{}
\titleformat{\subsubsection}{\normalsize\scshape}{\thesubsubsection}{0.5em}{}

\usepackage{fancyhdr}
\fancypagestyle{plain}{
  \fancyhf{}
  \fancyfoot[C]{\small\thepage}
  
}

\begin{document}

\twocolumn[
\begin{center}
\rule{\linewidth}{1pt}

\vspace{1.4em}
{\Large\bfseries
Learning Time-Inhomogeneous Markov Dynamics in Financial Time Series via Neural Parameterization
}
\vspace{0.5em}

\rule{\linewidth}{1pt}

\vspace{0.9em}

{\bfseries Jan Rovirosa \quad Jesse Schmolze}\\
{\small University of Wisconsin -- Madison, Madison, WI, USA}
\end{center}

\vspace{1.0em}
]

\thispagestyle{plain}

\begin{center}
\begin{minipage}{0.85\linewidth}
{\centering\bfseries Abstract\par}
\vspace{0.5em}
\small\normalfont
\justifying
\setlength{\parindent}{0em}
\setlength{\parskip}{0pt}

Modeling the dynamics of non-stationary stochastic systems requires balancing the representational power of deep learning with the mathematical transparency of classical models. While classical Markov transition operators provide explicit, theoretically grounded rules for system evolution, their empirical estimation collapses due to severe data sparsity when applied to high-resolution, high-noise environments. We explore this statistical barrier using financial time series as a canonical, real-world testbed. To overcome the degeneracy of empirical counting, we introduce a framework that utilizes neural networks strictly as parameterization engines to generate explicit, time-varying Markov transition matrices. By constraining the neural network to output its predictions as a formal stochastic operator, we maintain complete structural interpretability. We demonstrate that these learned operators successfully capture complex regime shifts: the state-conditioned model achieves mean row heterogeneity $\bar{\rho} = 0.0073$ while the state-free ablation collapses to exactly zero, and operator row entropy correlates with realized variance at $r = -0.62$ ($p \approx 10^{-251}$), revealing that high-volatility regimes homogenize transition dynamics rather than diversify them. Furthermore, rather than enforcing the Chapman--Kolmogorov equations as a rigid structural requirement, we repurpose them as a localized diagnostic tool to pinpoint specific temporal windows where first-order memory assumptions break down. Ultimately, this framework demonstrates how neural networks can be constrained to make rigorous, classical operator analysis viable for complex real-world time series.

\end{minipage}
\end{center}

\vspace{0.75em}

\section{Introduction}
Learning the dynamics of non-stationary stochastic systems is a central challenge in machine learning. We are often caught between two extremes: interpretable probabilistic models (like Markov chains), which offer clear structural insights but struggle to adapt to constantly evolving environments, and modern neural sequence models, which handle complex dependencies with ease but hide their logic inside black-box latent representations. This paper proposes a middle ground: a framework that uses the flexibility of neural networks to model the analytical structures of classical stochastic processes in real-time. 

Financial returns provide a high-stakes setting for this challenge. Market dynamics are noisy, heavy-tailed, and subject to abrupt regime changes. A standard way to study these shifts is to discretize returns into a finite state space $\mathcal{S}=\{s_1, \dots, s_n\}$ and model the transitions using a matrix $A \in [0, 1]^{n \times n}$. However, a conflict arises when we seek high resolution. To capture the nuance of heavy-tailed returns, we require a fine discretization (a large $n$). But as the number of bins increases, the number of possible transitions grows quadratically ($n^2$). In the finite, noisy window of financial data, most of these $n^2$ transitions are never observed. Standard empirical counting (the bedrock of classical Markov modeling) breaks here. It produces degenerate operators, matrices filled with zeros and high-variance noise that fail to represent the true underlying dynamics and become useless for actual analysis.

We address this degeneracy by replacing counting with neural parameterization. Instead of tabulating how many times $s_i$ moved to $s_j$, we train a neural network to learn the underlying mapping from current features $F_t$ and states $X_t$ to a probability distribution over the next state. Mathematically, we model the system as a feature-conditioned, time-inhomogeneous Markov chain. At each time $t$, the dynamics are governed by a transition operator $\widehat{A}_t$ constructed row-by-row:

\begin{equation}
\widehat{A}_t(i, \cdot) = \hat{p}_\theta(\cdot \mid X_t = s_i, F_t)
\end{equation}
By evaluating the network for every possible $s_i \in \mathcal{S}$, we reconstruct a full, smooth transition matrix for that specific moment in time. This approach transforms transition estimation into a conditional density estimation problem. The neural network creates the framework, filling in the sparse entries of the matrix by interpolating patterns across similar states and features. 

The primary advantage of this framework is that it yields an object that is both flexible and mathematically interpretable. Because the output is a standard stochastic operator, we can apply classical diagnostic tools that are unavailable to generic deep learning models: we can plot transition heatmaps to see how "rules" shift during market crashes vs. calm periods, we can measure row heterogeneity (how much the current state actually matters) and dispersion (how uncertain the market is), and we can use the Chapman–Kolmogorov equation to check if the model’s long-term forecasts are logically consistent with its short-term steps, providing a sanity check for the model's internal logic.

\textbf{Contributions}:
\begin{itemize}
    \item We introduce a neural parameterization for time-inhomogeneous Markov operators that addresses the sparsity problem of fine discretization while remaining fully interpretable.

    \item We show that state conditioning introduces genuine operator structure: the state-conditioned model achieves mean row heterogeneity $\bar{\rho} = 0.0073$ across the full time series, while the state-free ablation collapses to exactly $\rho = 0.0000$ at every timestep by construction.

    \item We demonstrate that the learned operators track market regimes: row entropy correlates with realized variance at $r = -0.62$ ($p \approx 10^{-251}$), with high-volatility periods homogenizing transition rows rather than diversifying them.

    \item We provide a suite of operator-level diagnostics, including Chapman--Kolmogorov consistency checks that localize in time the windows where first-order Markov closure is most strained.
\end{itemize}

\section{Related Work}\label{sec:related_work}

\paragraph{Classical Markov structure in finance.}
Markov transition operators have a long history as interpretable
models of regime change in economics and finance. Hamilton's
Markov-switching framework~\cite{hamilton1989} showed that
nonstationary macroeconomic dynamics can be captured through
regime-dependent evolution governed by a discrete-state Markov
law, making transitions between economic phases analytically
tractable. In credit-risk settings, Lando and Skodeberg~\cite{lando2002}
study rating migration using continuous-time observations, showing
that transition estimation and departures from simple Markov
assumptions are themselves core empirical issues. More broadly,
financial returns are known to exhibit heavy tails, volatility
clustering, and other stylized facts~\cite{cont2001} that motivate
regime-sensitive probabilistic models while simultaneously making
fine-grained empirical estimation noisy and statistically fragile.
Mettle et al.~\cite{mettle2022} analyze exchange rates using
time-inhomogeneous finite-state Markov chains, illustrating the
value of allowing transition structure to evolve across market
conditions rather than assuming a single stationary operator.
Our work shares this motivation but departs from all of the above
in a key respect: rather than estimating operators from counts or
closed-form conjugate updates, we replace tabulation entirely with
a neural parameterization that remains tractable precisely where
empirical counting collapses.

\paragraph{Neural probabilistic modeling for time series.}
From an estimation perspective, our task is a conditional density
estimation problem over a finite state space. Under linear
parameterizations, this reduces to standard multinomial models
for which convex optimization provides classical algorithmic
guarantees~\cite{boyd2004}. Mixture density networks~\cite{bishop1994}
generalize this to neural parameterizations of full conditional
output distributions rather than point estimates, and modern
forecasting architectures such as DeepAR~\cite{salinas2020} learn
time-varying predictive distributions with autoregressive neural
sequence models. As surveyed in~\cite{lim2021}, probabilistic
output heads are now standard in deep time-series forecasting.
The critical distinction between these approaches and ours is
structural: existing neural forecasters represent dynamics through
latent recurrent states or distribution parameters at the
observation level. They do not produce an explicit stochastic
operator on a fixed state space that can be directly inspected,
composed across time steps, or subjected to operator-theoretic
diagnostics such as the Dobrushin coefficient or Chapman--Kolmogorov
consistency checks.

\paragraph{Input-conditioned and time-varying Markov models.}
Bengio and Frasconi's input-output HMM~\cite{bengio1995} provides
an important precursor to our framework: latent-state transitions
are conditioned on an observed input sequence, linking classical
Markov structure with learned, input-dependent dynamics. Our
setting differs in two respects. First, the state of interest is
not latent but is an explicit discretization of the observed
return process. Second, the primary object we wish to estimate
and inspect is the induced transition operator itself, not the
latent state sequence. This distinction matters because our goal
is structural analysis of how transition rules vary over time, not
merely sequential prediction.

\paragraph{Neural--Markov hybrids and operator-first modeling.}
Several works combine Markov ideas with neural architectures.
Awiszus and Rosenhahn~\cite{awiszus2018} construct neural networks
capable of simulating Markov-chain behavior in stochastic
generative settings. Bellemare et al.~\cite{bellemare2017}, in
distributional reinforcement learning, emphasize that discrete
probability distributions should be treated as primary predictive
objects rather than collapsed to expectations, a perspective
broadly compatible with our emphasis on learning full transition
distributions directly. The central difference from both is that
our framework is operator-first: the network is used strictly as a
parameterization engine for an explicit, time-varying stochastic
operator whose rows remain valid probability distributions over a
fixed state space. This is precisely what allows classical operator
analysis (heatmap inspection, row heterogeneity, Dobrushin
contraction, Chapman--Kolmogorov diagnostics) to survive in a
setting where direct empirical counting fails.

\section{Problem Formulation}\label{sec:problem_formulation}

Our objective is to model the evolution of a stochastic system whose transition rules change over time.
We represent the system as a feature-conditioned, time-inhomogeneous Markov model and focus on the induced transition operators.

\subsection{State Space and Discretization}
Let $\{P_t\}_{t\ge 0}$ denote adjusted daily closing prices and define the one-day return
\begin{equation}
r_t := \frac{P_t - P_{t-1}}{P_{t-1}}.
\end{equation}
To enable a structural analysis of regime dynamics, we discretize returns into a finite state space
$\mathcal{S}=\{s_1,\dots,s_n\}$ via a map $\phi:\mathbb{R}\to\mathcal{S}$, and define the Markov state
\begin{equation}
X_t := \phi(r_t)\in\mathcal{S}.
\end{equation}
We prioritize relatively fine discretizations (large $n$). While this provides a detailed view of the return distribution,
it introduces a sparsity problem: the number of potential state-to-state transitions scales as $n^2$, which can exceed the
available sample support, making empirical (count-based) transition estimators unreliable.

\subsection{Feature-Conditioned One-Step Dynamics}
Let $F_t\in\mathbb{R}^d$ denote observable covariates available at time $t$ (exogenous indicators and/or lagged features).
We assume a first-order Markov property conditional on $F_t$:
\begin{equation}
\begin{split}
\mathbb{P}(X_{t+1}=s_j \mid X_t=s_i,\; X_{t-1},X_{t-2},\dots,\; F_t) = \\
=\; \mathbb{P}(X_{t+1}=s_j \mid X_t=s_i,\; F_t).
\end{split}
\end{equation}
Unlike stationary Markov chains, the transition law is allowed to vary with time through its dependence on $F_t$.

\subsection{Two Operator Views and Prediction Targets}
Our experiments use two complementary label constructions, leading to two operator interpretations.

\paragraph{(i) Forward-return (distributional) targets.}
For horizon $h\ge 1$, define the forward return from $t+1$ to $t+1+h$ by
\begin{equation}
R_t^{(h)} := \frac{P_{t+1+h}-P_{t+1}}{P_{t+1}},
\end{equation}
and discretize it into $m_h$ bins to obtain $Y_t^{(h)}\in\mathcal{Y}^{(h)}$ with $|\mathcal{Y}^{(h)}|=m_h$.
This yields a (generally rectangular) population operator $A_t^{(h)}\in[0,1]^{n\times m_h}$ defined by
\begin{equation}
A_t^{(h)}(i,j) := \mathbb{P}(Y_t^{(h)}=j \mid X_t=s_i,\; F_t).
\end{equation}

\paragraph{(ii) State-to-state targets (operator diagnostics).}
For operator diagnostics (including Chapman--Kolmogorov checks), we instead set
\begin{equation}
Y_t^{(h)} := X_{t+h}\in\mathcal{S},
\end{equation}
which produces a square population operator $A_t^{(h)}\in[0,1]^{n\times n}$:
\begin{equation}
A_t^{(h)}(i,j) := \mathbb{P}(X_{t+h}=s_j \mid X_t=s_i,\; F_t).
\end{equation}
In this setting, multi-step composition is well-defined on the common state space $\mathcal{S}$.

\subsection{Neural Parameterization: a Row-wise Operator Estimator}
To avoid unstable counting in high-dimensional discretizations, we estimate transition rows using a neural model
$f_\theta^{(h)}$. For a given state $s_i$ and features $F_t$, the model outputs a probability vector over the label space:
\begin{equation}
\widehat{A}_t^{(h)}(i,\cdot) := f_\theta^{(h)}(s_i, F_t)\in\Delta^{m_h-1},
\end{equation}
where $\Delta^{m_h-1}$ is the probability simplex (and $m_h=n$ in the state-to-state diagnostic setting).
To construct the full estimated operator at time $t$, we fix $F_t$ and evaluate the network for each $s_i\in\mathcal{S}$:
\begin{equation}
\text{for } i=1,\dots,n:\quad \widehat{A}_t^{(h)}(i,\cdot)\leftarrow f_\theta^{(h)}(s_i,F_t).
\end{equation}
This replaces sparse, noisy transition counts with a shared-parameter estimator that produces well-defined operator rows even when particular transitions are rarely observed.

\section{Methods and Models}\label{sec:methods_and_models}

\subsection{Data, Alignment, and Preprocessing}

Our empirical study focuses on a single equity (JPM) together with a collection of exogenous covariates obtained from public and commercial financial data sources. The covariates comprise (i) daily market and macroeconomic indicators (e.g., policy-rate and credit-spread proxies) and (ii) lower-frequency firm fundamentals (e.g., quarterly accounting variables and ratios). Because many features are reported at lower frequency than prices, all covariates are aligned to the daily grid by carrying forward the most recently observed value until the next release date; short internal gaps are filled by linear interpolation, while features requiring extensive extrapolation are excluded.

Each feature is standardized using training-set statistics only, producing a normalized feature vector $F_t\in\mathbb{R}^d$. When using reduced feature sets, we rank features by mutual information with the training labels (computed on the training split only) and retain the top-$k$ features.

\subsection{Discretization Choices}
We discretize one-day returns into $n=55$ states using quantile binning fit on the training segment. For horizon-$h$ forward-return labels, we discretize $R_t^{(h)}$ into $m_h$ bins (typically $m_h\in\{10,20,35,55\}$). For operator diagnostics we set $Y_t^{(h)}=X_{t+h}$, which yields square operators on $\mathcal{S}$.

\subsection{Neural Operator Parameterization}
For a given horizon $h$, the model takes as input the concatenation
of a one-hot state encoding $e(X_t) \in \mathbb{R}^n$ and the
feature vector $F_t \in \mathbb{R}^d$, and passes it through an
MLP $g^{(h)}_\theta$ followed by a softmax to produce a valid
probability distribution over the $m_h$ output bins:
\begin{equation}
\widehat{A}^{(h)}_t(X_t,\cdot)
\;=\;
\mathrm{softmax}\!\left(
  g^{(h)}_\theta\!\left([\,e(X_t)\,;\,F_t\,]\right)
\right)
\in \Delta^{m_h - 1}.
\label{eq:neural_operator_row}
\end{equation}
Evaluating this for every $s_i \in \mathcal{S}$ under the same
$F_t$ assembles the full operator $\widehat{A}^{(h)}_t \in
[0,1]^{n \times m_h}$, which is square and stochastic when
$m_h = n$.

\subsection{Optimization and Regularization}
Models are trained using Adam with early stopping based on validation negative log-likelihood. 
To regularize learning under noisy discretization, we optionally employ a smoothed-target objective that replaces the one-hot label $Y_t^{(h)}$ with a distribution $\tilde{q}_t^{(h)} \in \Delta^{m_h-1}$ allocating partial mass to neighboring bins:
\begin{equation}\label{eq:smoothed_ce_v2}
\mathcal{L}^{(h)}_{\mathrm{smooth}}(\theta)
= -\sum_{t \in \mathcal{T}_{\mathrm{train}}}
  \sum_{j=1}^{m_h} \tilde{q}^{(h)}_{t,j}
  \log \hat{p}^{(h)}_t(j).
\end{equation}
All preprocessing steps (bin-edge fitting, feature standardization, feature ranking) are performed using training data only. 
Reported results use chronological train/validation/test splits rather than random i.i.d.\ splits.

\subsection{State-Free Ablation Baseline}
To isolate the contribution of explicit Markov state conditioning, we also consider a state-free baseline that removes $e(X_t)$ from the input:
\begin{equation}
\hat{p}^{(h)}_{\theta,\mathrm{sf}}(\cdot\mid F_t)
=
\mathrm{softmax}\!\left(g^{(h)}_{\theta,\mathrm{sf}}(F_t)\right).
\end{equation}
In the state-to-state setting ($m_h=n$), this baseline corresponds to an operator whose rows are identical at each time $t$, serving as a reference for diagnosing the extent to which learned dynamics depend on the Markov state.

\section{Experiments}\label{sec:experiments}

To validate our neural parameterization framework, we design an experimental suite that explicitly tests the limits of classical transition estimation and evaluates the structural interpretability of the learned operators. 

\subsection{Dataset, Horizons, and Splits}
We evaluate the framework on a single-equity case study (JPM). The dataset consists of $T$ daily trading observations, each with an aligned feature vector $F_t\in\mathbb{R}^d$. To strictly avoid look-ahead bias, all evaluations utilize chronological splits. For each look-ahead horizon $h \in \{1,2,5,10\}$, we construct a horizon-specific dataset and apply a fixed train/validation/test split of $70\%/15\%/15\%$ along the time axis.

The Markov state space is fixed to a fine discretization of $n=55$ bins ($X_t\in\mathcal{S}$), fit via quantiles on the training segment to ensure approximately balanced marginal state visitation. 

\subsection{Prediction Targets}
As formulated in Section~\ref{sec:problem_formulation}, we use
two label constructions: forward-return targets (rectangular
operators, $m_h \in \{10, 20, 35, 55\}$) for assessing predictive
dynamics, and state-to-state targets ($Y_t^{(h)} := X_{t+h}$,
square operators) for structural diagnostics and CK composition.

\subsection{Models and Baselines}
All neural models share an identical base Multi-Layer Perceptron (MLP) architecture and differ solely in their inductive biases regarding the Markov state.

\paragraph{State-Conditioned Transition Model.}
Our primary model, which parameterizes the operator row $\widehat{A}_t^{(h)}(X_t,\cdot)$ via $[e(X_t); F_t]$. By explicitly conditioning on the current state $X_t$, this model acts as a dynamic operator generator.

\paragraph{State-Free Ablation Baseline.}
To isolate the value of the Markov inductive bias, we train a baseline conditioned only on features ($\hat{p}_{\theta,\mathrm{sf}}^{(h)}(\cdot\mid F_t)$). By definition, this model outputs an operator with identical rows at each time step (row-invariant). It serves as a negative control: if the state-conditioned model collapses to look like this baseline, the Markov state carries no useful transition information.

\paragraph{Count-Based Estimators.}
To represent the classical approach, we evaluate empirical estimators derived from training counts of $(X_t, Y_t^{(h)})$. These include a marginal baseline, a direct conditional estimator, and a backoff-smoothed estimator (with validation-tuned hyperparameters). These baselines are included specifically to demonstrate the degeneracy of empirical counting under fine discretization.

\subsection{Training Protocol}
The shared neural architecture is an MLP with hidden widths $64 \rightarrow 128 \rightarrow 256 \rightarrow 128 \rightarrow 64$, using GELU activations and dropout ($p=0.2$). Models are optimized via Adam with weight decay and gradient clipping. Early stopping is governed by validation negative log-likelihood. To account for the noise inherent in discretized returns, we test both hard cross-entropy and a smoothed-label objective that allocates partial mass to neighboring bins. 

\subsection{Evaluation Metrics (Operator-Focused)}
In alignment with our framework, our evaluation prioritizes operator structure and diagnostic interpretability over raw point prediction.

\paragraph{Degeneracy and Sparsity.}
To quantify the failure modes of classical methods, we measure the sparsity of the empirical count matrices $C\in\mathbb{N}^{n\times m_h}$ on the training data. Metrics include the fraction of absolute zero cells and the fraction of cells falling below a minimum observation threshold (e.g., $<5$ samples). 

\paragraph{Operator Interpretability Diagnostics.}
For the square one-step operators $\widehat{A}_t^{(1)}$, we compute time-varying structural statistics:
\begin{itemize}
    \item \textit{Row Heterogeneity:} The average pairwise total-variation (TV) distance between rows, acting as a direct, mathematical measure of state dependence.
    \item \textit{Row Entropy:} The mean entropy across rows, measuring the operator's forward dispersion.
    \item \textit{Dobrushin Coefficient:} A measure of the operator's contractive properties (row separation in $\ell_1$).
\end{itemize}

\paragraph{Chapman--Kolmogorov (CK) Consistency.}
In the state-to-state setting, we measure the divergence between the directly estimated $h$-step operator $\widehat{A}_t^{(h)}$ and the composed product $\widehat{A}_t^{(1:h)}$. This is quantified using row-wise KL divergence and TV distance, localized over time to identify regime windows where the first-order Markov assumption weakens.

\paragraph{Sanity Checks.}
Finally, to ensure the learned operators reflect valid probability distributions, we compute the test-set Negative Log-Likelihood (NLL) relative to a marginal baseline. We assess calibration via Expected Calibration Error (ECE) for coarse regimes like negative returns, and report robustness through block bootstrapping over time to compute confidence intervals on held-out log-likelihood.

\section{Results}\label{sec:results}

The results follow a cumulative structural argument: classical
estimation fails at our target resolution, neural parameterization
produces inspectable operators, and those operators carry
quantifiable state-dependent structure that tracks market regimes and
supports a localized Chapman--Kolmogorov diagnostic. Predictive
validation is reported in full but is secondary to the structural
findings.

%-------------------------------------------------------------------
\subsection{The Sparsity Barrier: Why Empirical Estimation Fails}
\label{sec:results_degeneracy}

\begin{figure}[htpb]
    \centering
    \includegraphics[width=0.95\linewidth]{%
        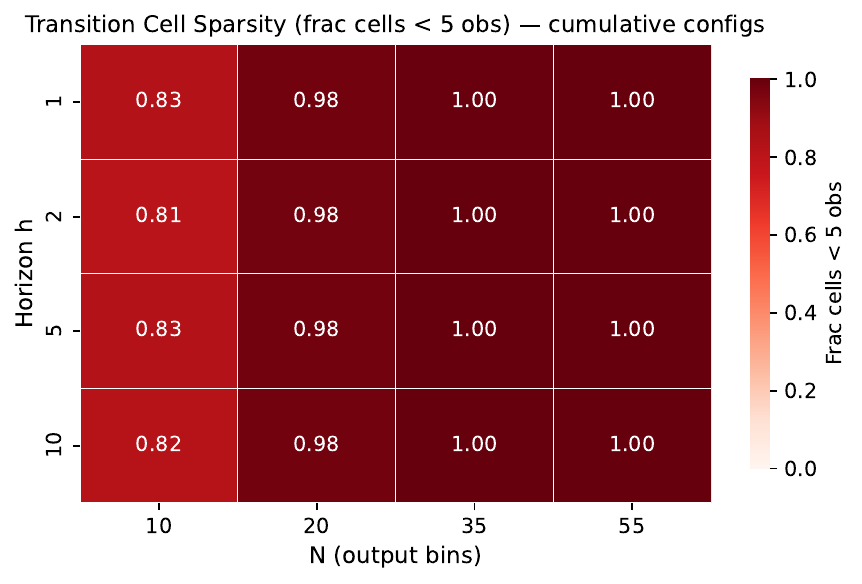}
    \caption{Empirical transition degeneracy for forward-return
    tasks. Each cell reports the fraction of entries in $C^{(h)}$
    with fewer than 5 observations. At $N = 35$ and $N = 55$, the
    fraction is identically 1 across all horizons.}
    \label{fig:transition_sparsity_heatmap}
\end{figure}

For each configuration $(h, m_h)$ in the forward-return setting, we
construct the empirical count matrix
$C^{(h)} \in \mathbb{N}^{n\times m_h}$ on the training split:
\begin{equation}\label{eq:empirical_count}
C^{(h)}(i,j) := \#\{t \in \mathcal{T}_{\mathrm{train}}
  : X_t = s_i,\; Y_t^{(h)} = j\}.
\end{equation}
Although marginal states $X_t$ are quantile-binned to ensure roughly
equal visitation, the \emph{joint} transition structure collapses
once we condition on the current state. At $m_h = N = 55$ bins,
$99.9\%$ of transition cells contain fewer than five observations
at every horizon $h \in \{1, 2, 5, 10\}$, and over $57\%$ are
entirely unobserved. The median row has nonzero entries in only
23 of 55 possible output bins. This collapse is horizon-invariant:
the degeneracy profile is nearly identical across all four horizons
(Figure~\ref{fig:transition_sparsity_heatmap}),
ruling out the possibility that longer look-ahead accumulates enough
joint support to rescue the empirical estimator. Direct counting
cannot serve as the basis for operator-level analysis at this
discretization.

\FloatBarrier

%-------------------------------------------------------------------
\subsection{Inspectable Transition Operators}
\label{sec:results_operators}

The first payoff of neural parameterization is that, for any time
$t$, we can produce a complete stochastic operator
$\widehat{A}_t^{(1)} \in [0,1]^{n\times n}$ by evaluating the
network at each input state $s_i$ under the same feature snapshot
$F_t$. Unlike generic sequence models, whose predictions are
mediated by opaque latent states, the output is a probability matrix
that can be directly inspected.

\begin{figure}[htpb]
    \centering
    \includegraphics[width=0.8\linewidth]{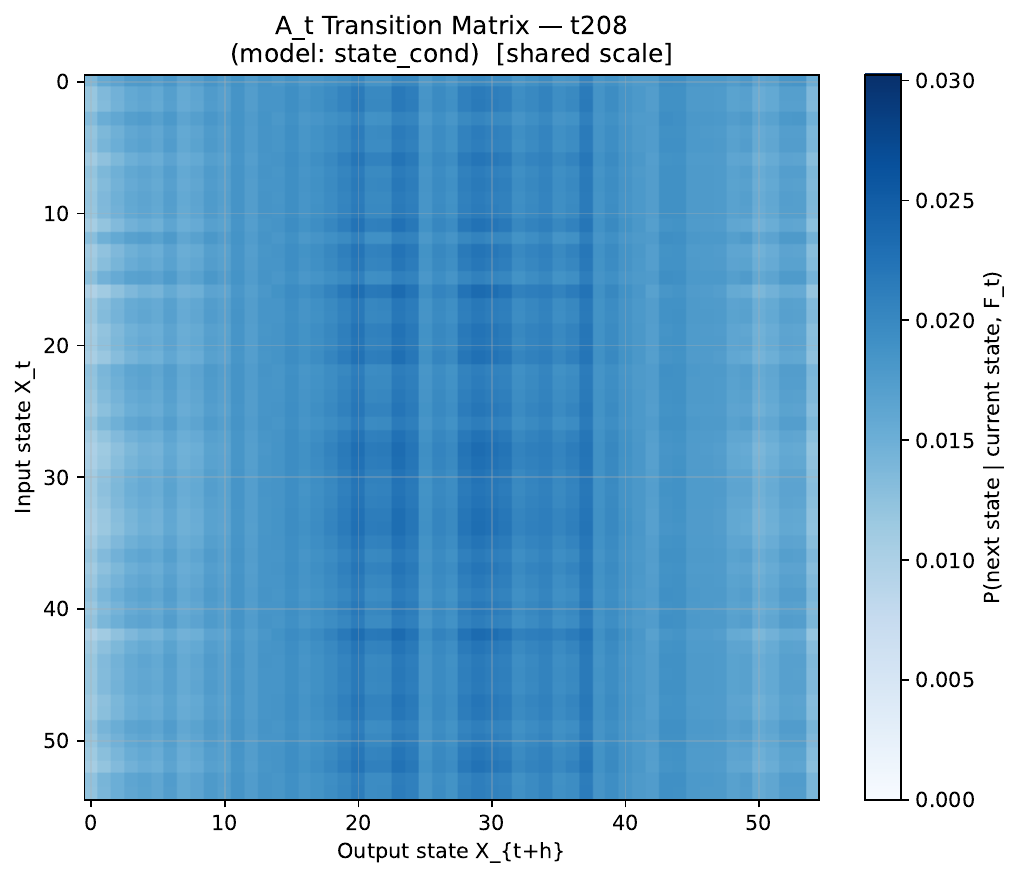}
    \caption{Snapshot of $\widehat{A}_t^{(1)}$ at $t=208$ (tranquil regime),
    state-conditioned model.}
    \label{fig:At_cond_t208}
\end{figure}

\begin{figure}[htpb]
    \centering
    \includegraphics[width=0.8\linewidth]{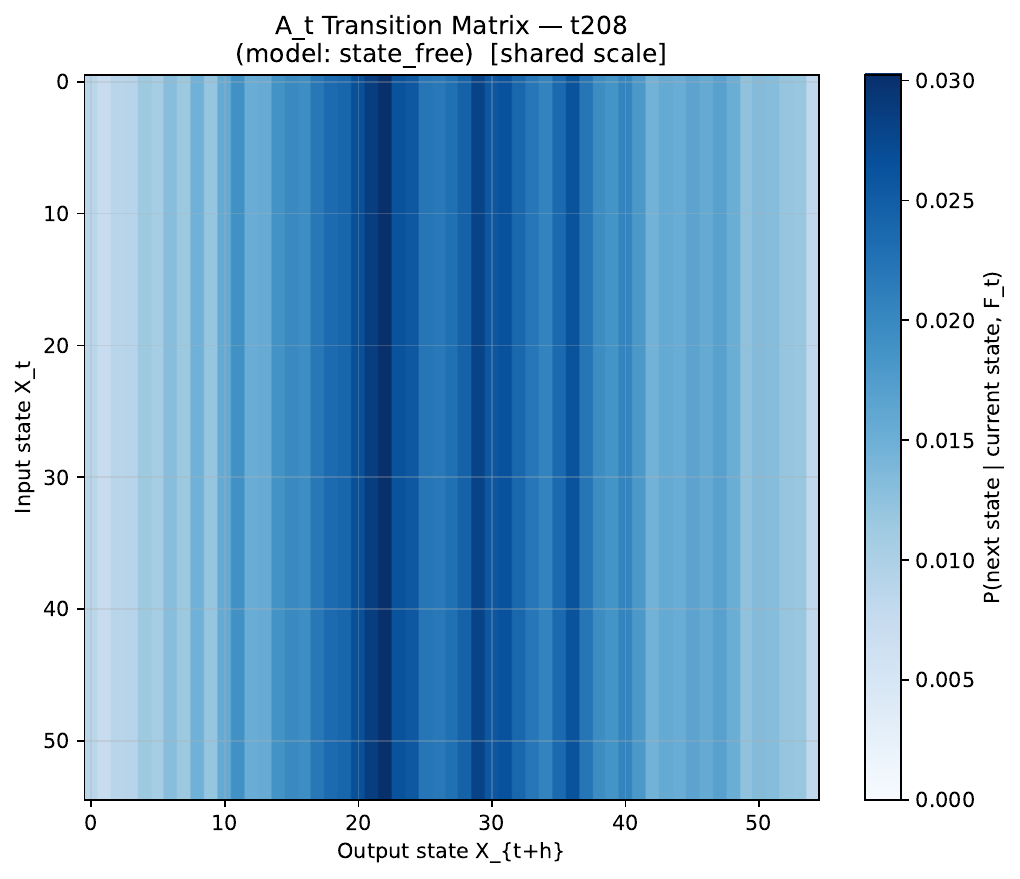}
    \caption{Snapshot of $\widehat{A}_t^{(1)}$ at $t=208$ (tranquil regime),
    state-free baseline. Row-invariant by construction.}
    \label{fig:At_free_t208}
\end{figure}

\begin{figure}[htpb]
    \centering
    \includegraphics[width=0.8\linewidth]{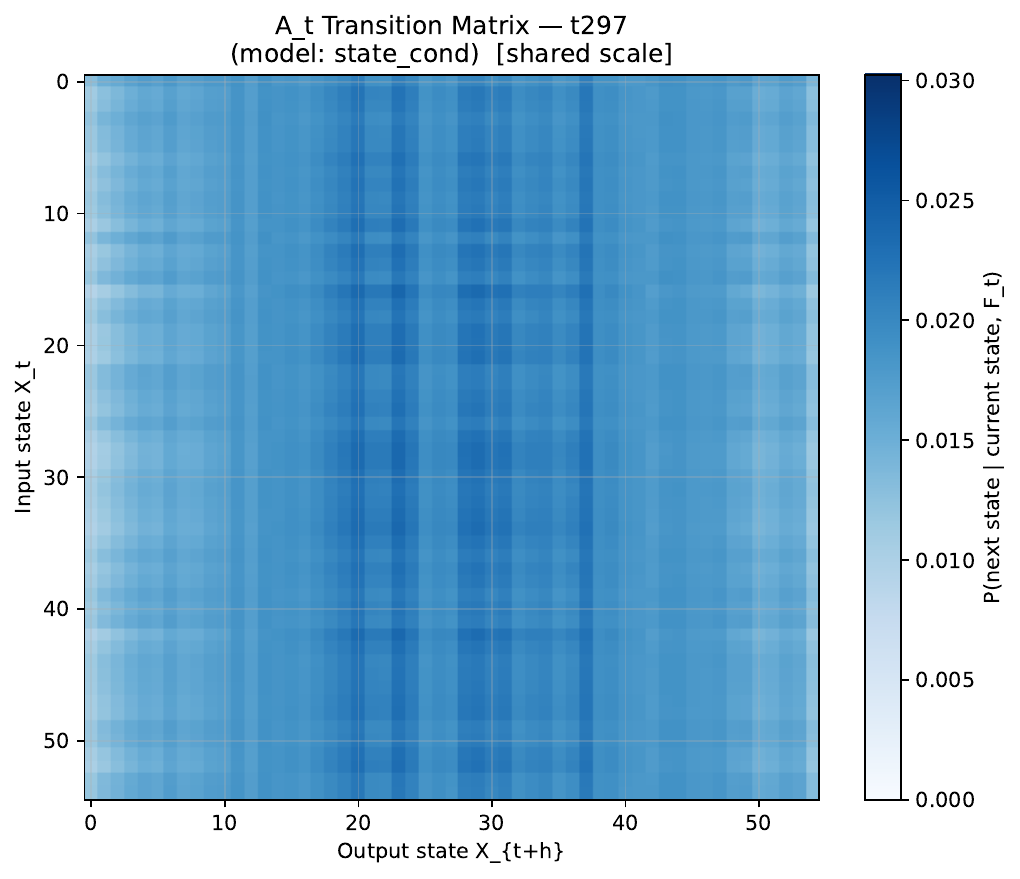}
    \caption{Snapshot of $\widehat{A}_t^{(1)}$ at $t=297$ (stressed regime),
    state-conditioned model. }
    \label{fig:At_cond_t297}
\end{figure}

\begin{figure}[htpb]
    \centering
    \includegraphics[width=0.8\linewidth]{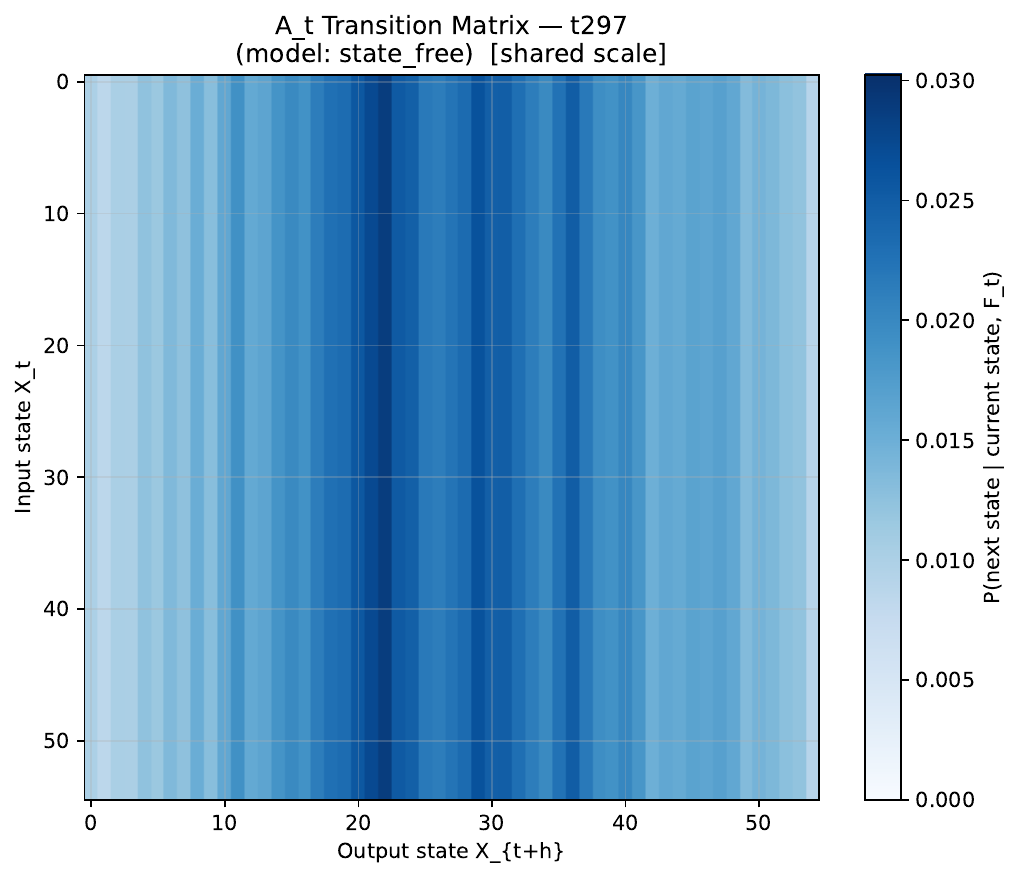}
    \caption{Snapshot of $\widehat{A}_t^{(1)}$ at $t=297$ (stressed regime),
    state-free baseline. Row-invariant by construction.}
    \label{fig:At_free_t297}
\end{figure}

Figures~\ref{fig:At_cond_t208}, ~\ref{fig:At_free_t208}, ~\ref{fig:At_cond_t297} and ~\ref{fig:At_free_t297} show snapshots at two selected timesteps. The state-conditioned model produces complex,
diagonal-heavy structures whose rows vary substantially across
states, capturing the persistence and state-dependent dispersion of
the return process. The state-free baseline, by construction,
produces row-invariant operators (horizontal stripes): the same
next-step distribution is assigned regardless of the current state.
This visual contrast motivates the quantitative analysis that
follows.

\subsection{Operator Diagnostics and Regime Structure}
\label{sec:results_diagnostics}

Following the evaluation metrics defined in
Section~\ref{sec:experiments}, we compute three diagnostics from
$\widehat{A}_t^{(1)}$ at every timestep over the full time series:
row heterogeneity $\rho(\widehat{A}_t)$, row entropy
$H(\widehat{A}_t)$, and the Dobrushin coefficient
$\delta(\widehat{A}_t)$.

\begin{figure}[htbp]
  \centering
  \includegraphics[width=\linewidth]{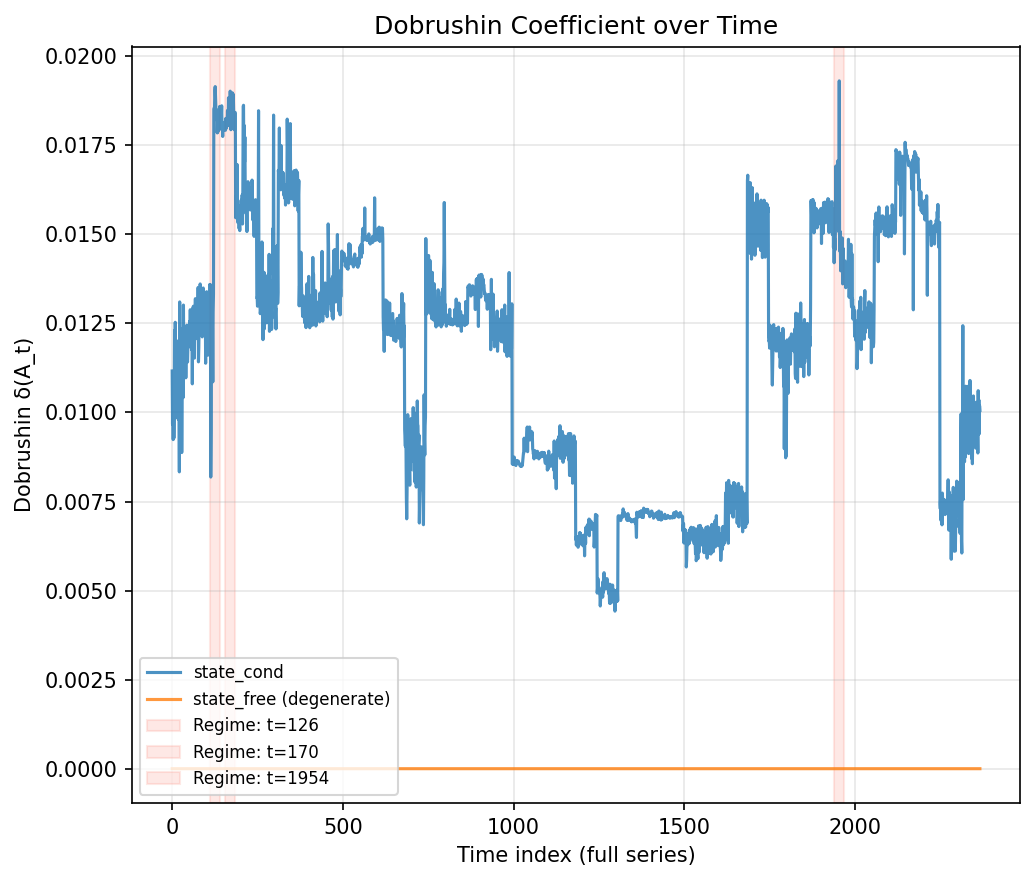}
  \caption{Dobrushin coefficient $\delta(\widehat{A}_t)$ over the full
time series for the state-conditioned model (blue) and state-free
baseline (orange). The baseline collapses to zero by construction.
Shaded regions indicate selected high-volatility windows.}
  \label{fig:dobrushin}
\end{figure}

\paragraph{State conditioning introduces genuine structure.}
Over the full series, the state-conditioned model exhibits mean row
heterogeneity $\bar{\rho} = 0.0073$ (std $0.0021$, peak $0.014$),
while the state-free baseline produces $\rho = 0.0000$ at every
timestep to machine precision, confirming that
the Markov state $X_t$ is informative about which row of the
operator to apply. The Dobrushin coefficient averages $0.029$ (peak
$0.057$), and row entropy averages $3.93$ nats against the uniform
upper bound $\log 55 \approx 4.01$, placing the operators near
maximum entropy on average yet carrying measurable state-dependent
structure, consistent with the low signal-to-noise ratio of daily
returns. Figure~\ref{fig:dobrushin} tracks all three
diagnostics over time.

We note that at $h = 2$ and $h = 10$, the directly trained
state-conditioned operators are near-uniform (value range
$[0.016, 0.020] \approx 1/55$), indicating that little
state-dependent structure is recovered at those horizons given the
available data. At $h = 5$, the operators are genuinely
concentrated (maximum entry $0.110$), and we accordingly anchor the
multi-step analysis of Section~\ref{sec:results_ck} at $h = 5$.

\begin{figure}[htpb]
    \centering
    \includegraphics[width=1\linewidth]{%
        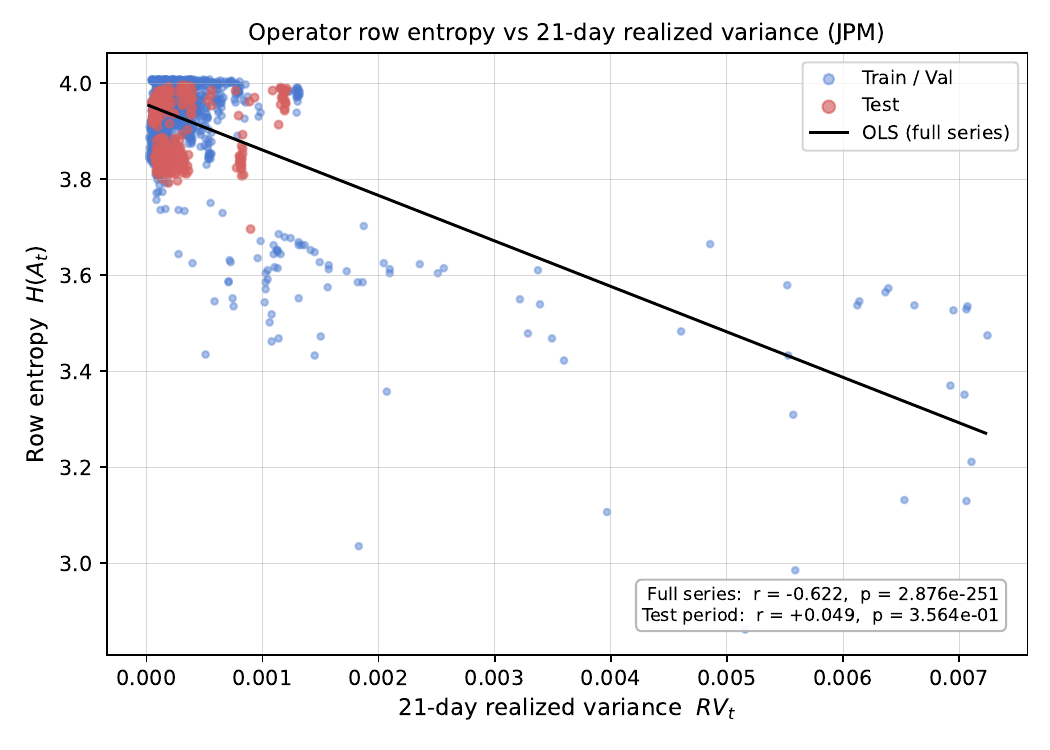}
    \caption{Row entropy $H(\widehat{A}_t)$ versus 21-day realized
    variance $\mathrm{RV}_t$ computed from JPM daily returns. Each
    point is one trading day; colors indicate chronological split.
    The full-series Pearson correlation is $r = -0.622$
    ($p \approx 10^{-251}$, $T = 2{,}347$): higher realized variance
    is strongly associated with more concentrated transition
    operators. The test period occupies a narrow $\mathrm{RV}_t$
    range.}
    \label{fig:entropy_vs_rv}
\end{figure}

\paragraph{High volatility homogenizes the operator.}
Aligning the diagnostic time series with a rolling 21-day realized
variance proxy $\mathrm{RV}_t$ computed directly from $r_t$ reveals
that high-volatility regimes \emph{homogenize} the operator rather
than diversify it. Stratifying timesteps into the top and bottom
$20\%$ of $\mathrm{RV}_t$, the mean row heterogeneity drops from
$0.0103$ in tranquil regimes to $0.0044$ in stressed ones (a
factor of $2.3\times$ difference). Both row entropy and row
heterogeneity differ significantly across regimes (two-sample
$t$-test, $p < 0.02$ for every metric). The full-series Pearson
correlation between $H(\widehat{A}_t)$ and $\mathrm{RV}_t$ is
$r = -0.62$ ($p \approx 10^{-251}$, $T = 2{,}347$), visualized in
Figure~\ref{fig:entropy_vs_rv}. During market stress, transition
rows collapse toward a similar, concentrated distribution: volatile
periods reduce the diversity of next-state dynamics regardless of
the current state. The same correlation computed on the held-out
test period alone is not statistically significant
($r = +0.049$, $p = 0.36$); as Figure~\ref{fig:entropy_vs_rv}
shows, the test period occupies a narrow range of realized variance
insufficient to identify the relationship.

\begin{figure}[htbp]
  \centering
  \includegraphics[width=\linewidth]{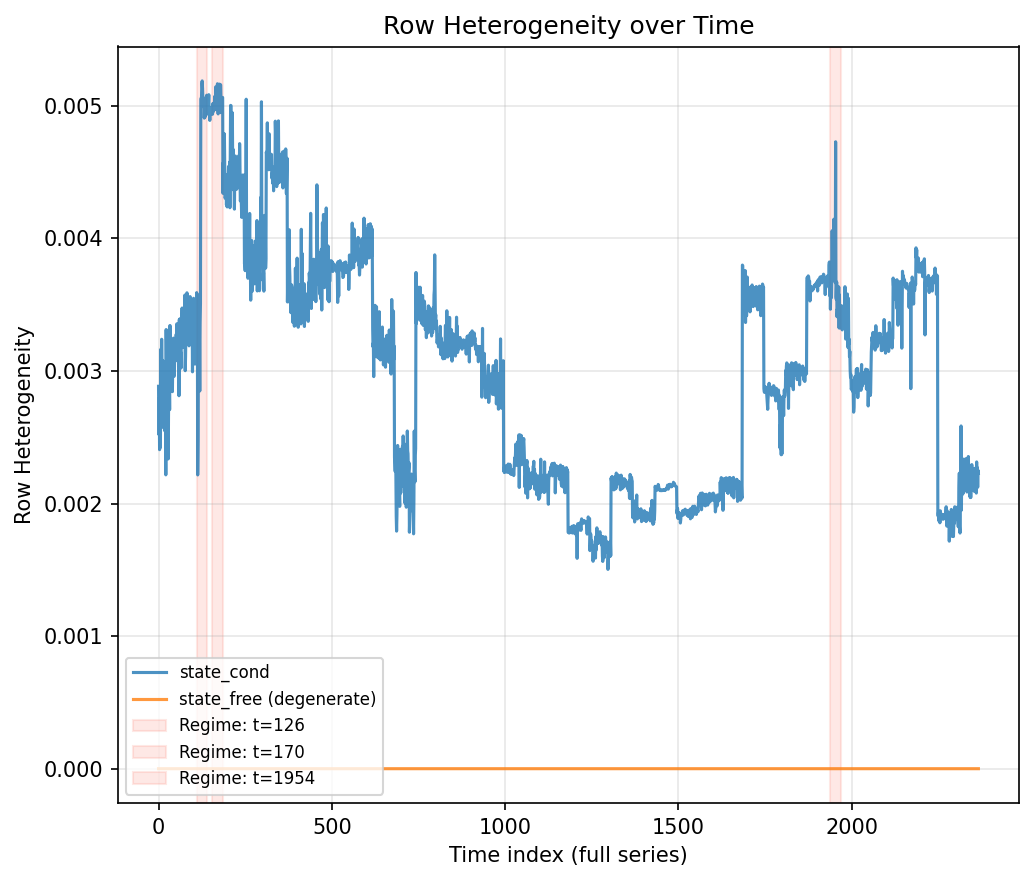}
  \caption{Mean pairwise TV distance between rows (row heterogeneity
$\rho(\widehat{A}_t)$) over the full time series. The state-free
baseline is identically zero; the state-conditioned model exhibits
meaningful time variation, with elevated values during tranquil
regimes.}
  \label{fig:row-heterogeneity}
\end{figure}

\begin{figure}[htbp]
  \centering
  \includegraphics[width=\linewidth]{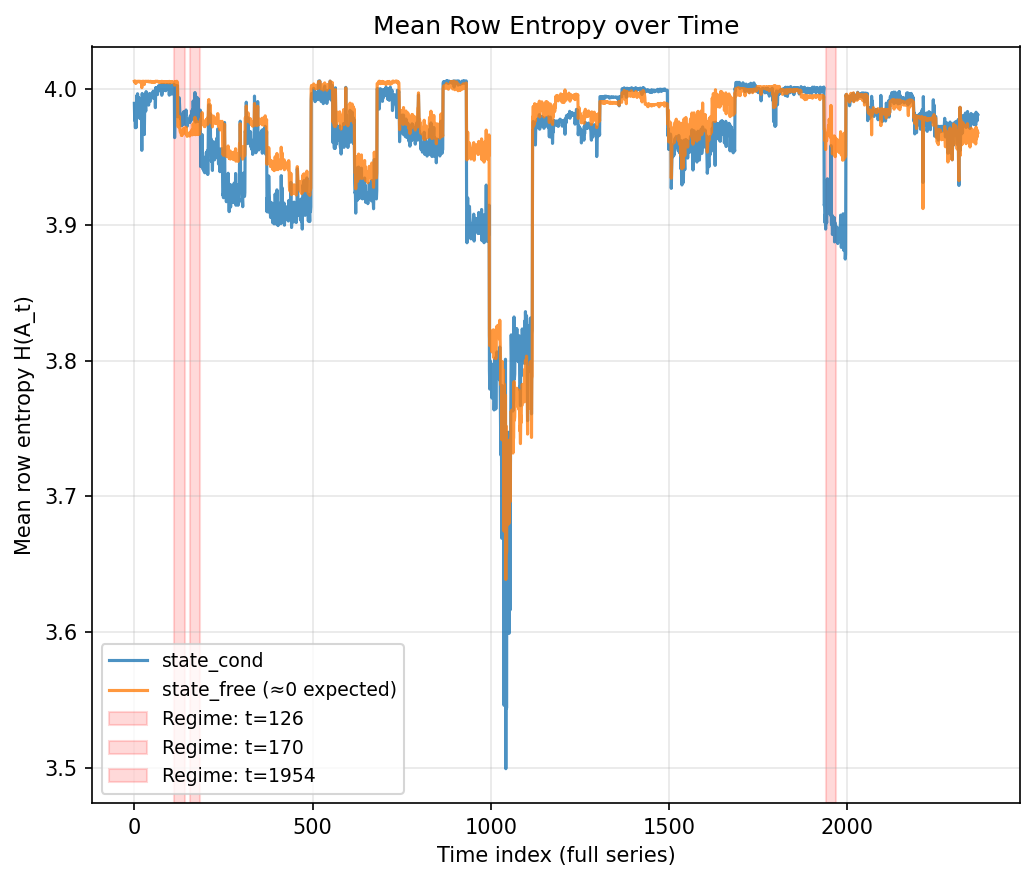}
  \caption{Mean row entropy $H(\widehat{A}_t)$ over the full time
series. Entropy declines sharply during the COVID-19 crash window
($t \approx 1031$--$1114$), where the model concentrates
probability mass on extreme return bins consistent with observed
bimodal crisis dynamics.}
  \label{fig:row-entropy}
\end{figure}

\FloatBarrier

%-------------------------------------------------------------------
\subsection{The Chapman--Kolmogorov Diagnostic}
\label{sec:results_ck}

Because the model outputs stochastic operators on a shared state
space, we can probe their internal structural consistency using the
time-inhomogeneous Chapman--Kolmogorov (CK) equations. For horizon
$h$ and timestep $t$, we compare the direct $h$-step prediction
$\widehat{A}_t^{(h)}$ to the composition
\begin{equation}
    \widehat{A}_t^{(1:h)} :=
    \widehat{A}_t^{(1)}\,\widehat{A}_{t+1}^{(1)} \cdots
    \widehat{A}_{t+h-1}^{(1)},
\end{equation}
and report the row-averaged KL divergence. As noted in
Section~\ref{sec:results_diagnostics}, we anchor the analysis at
$h = 5$, where the directly trained operators carry genuine
state-dependent structure.

\paragraph{Aggregate discrepancy.}
Averaged over the test period at $h = 5$, the state-conditioned
model has mean CK KL divergence $0.140$, compared to $0.022$ for
the state-free baseline (roughly $6\times$ larger). A strict
first-order Markov closure in $X_t$ is only an approximation in
real return data; a perfectly zero CK error would be suspicious.
The larger discrepancy for the state-conditioned model is the
expected consequence of learning genuinely state-dependent dynamics
that the iterated one-step product cannot fully reconstruct.

\begin{figure}[htpb]
    \centering
    \includegraphics[width=0.98\linewidth]{%
        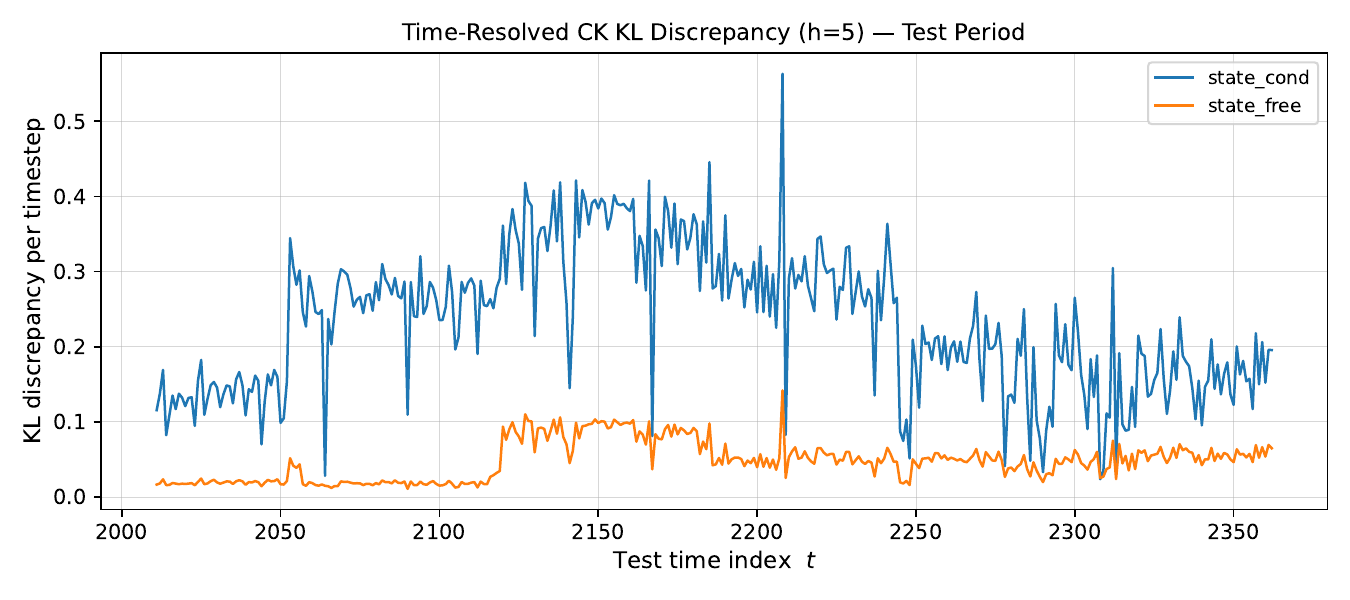}
    \caption{Time-resolved CK KL discrepancy at $h = 5$ over the
    test period. The state-conditioned model (blue) maintains a
    substantially higher and more variable discrepancy than the
    state-free baseline (orange), with a pronounced elevated band
    where operator diagnostics also indicate concentrated,
    low-entropy dynamics.}
    \label{fig:ck_error_time_series_h5}
\end{figure}

\paragraph{Temporal structure of CK violations.}
The more informative quantity is the per-timestep KL series.
Figure~\ref{fig:ck_error_time_series_h5} shows that CK violations
are non-uniform in time and concentrate in an extended band roughly
covering $t \in [2120, 2200]$, where the state-conditioned model's
discrepancy rises to $0.3$--$0.4$ while the state-free baseline
remains well below $0.15$. As an internal consistency check, the
per-timestep CK KL at $h = 5$ correlates with row entropy at
$r = -0.533$ ($p \approx 0$): two diagnostic lenses applied to the
same learned operators flag the same windows. This gives the CK
diagnostic its practical role in the framework: a localized probe
of where the first-order Markov approximation is most strained,
available only because we expose an explicit operator.

\FloatBarrier

%-------------------------------------------------------------------
\subsection{Predictive Validation and Calibration}
\label{sec:results_predictive}

While predictive accuracy is not the focus of this single-asset
study, we computed standard log-likelihood and calibration metrics
to confirm that the learned operators represent valid probability
distributions grounded in the data.
Table~\ref{tab:main_results_h10} reports test-set $\Delta$NLL relative
to a marginal baseline (positive = better than marginal) with
block-bootstrap $95\%$ confidence intervals at $N = 55$ across all
horizons. Across configurations at $N = 55$, the strongest result is
$\Delta\mathrm{NLL} = +0.025$ at $h = 10$ for the state-conditioned
model (95\% CI $[-0.020, +0.028]$), with the state-free model
reaching $+0.016$ at the same horizon. Shorter horizons are near
or below marginal (at $h = 1$, the state-conditioned model
records $\Delta\mathrm{NLL} = -0.033$ (CI
$[-0.085, -0.003]$)) consistent with weak-form efficiency and with
the limitations of the single-asset feature set described in
Section~\ref{sec:experiments} (many covariates are interpolated or
carried forward at low frequency).

For distributional validity, at $h = 1$, $N = 55$, the
state-conditioned model achieves Expected Calibration Error
$\mathrm{ECE} = 0.079$ on the negative-return event, compared to
$0.095$ for the state-free baseline and $0.081$ for the backoff
estimator (the best calibration of the three). 

These modest predictive gains are a natural consequence of the
study's scope: a single equity with daily granularity and a
low-frequency feature set provides limited signal-to-noise. An
ongoing extension of this framework to a multi-asset universe with
higher-frequency data and substantially larger sample sizes is
expected to improve predictive accuracy materially, while
preserving the operator-level interpretability that is the
contribution of the present work.

\begin{table}[htpb]
\centering
\caption{Test $\Delta$NLL at $h = 10$, $N = 55$ (positive =
better than marginal). Block-bootstrap 95\% CIs in brackets.
Held-out chronological test split.}
\label{tab:main_results_h10}
\small
\begin{tabular}{lcc}
\toprule
Model & $\Delta$NLL & 95\% CI \\
\midrule
Additive          & $+0.006$ & --- \\
Backoff           & $+0.007$ & --- \\
State-cond (ours) & $\mathbf{+0.025}$ & $[-0.020,\;+0.028]$ \\
State-free (ours) & $+0.016$ & $[-0.010,\;+0.036]$ \\
\bottomrule
\end{tabular}
\end{table}

\section{Discussion}\label{sec:discussion}

\subsection{Volatility and Operator Structure}
A concrete structural finding from the single-asset analysis concerns
the relationship between market volatility and operator geometry.
Rather than producing more heterogeneous rows under stress (as one
might expect if volatile markets presented more state-dependent
dynamics) the learned operators become more homogeneous during
high-volatility periods. Row entropy and row heterogeneity are both
significantly lower in the top-$20\%$ realized-variance windows than
in the bottom $20\%$ (two-sample $t$-test, $p < 0.02$), and the
full-series correlation between row entropy and realized variance
is $r = -0.62$ ($p \approx 10^{-251}$). The interpretation is that
volatile markets reduce the diversity of next-state dynamics
regardless of the current state: all transition rows collapse toward
a similar concentrated distribution under stress, while tranquil
periods admit richer, more diffuse row structures. We note that this
correlation is not statistically significant in the held-out test
period ($r = +0.049$, $p = 0.36$), where the range of realized
variance is too narrow to identify the relationship; we report both
values for transparency.

\subsection{Rethinking the Markov Assumption via CK Diagnostics}
A key conceptual shift in our framework is the repurposing of the Chapman--Kolmogorov (CK) equations. Rather than treating strict CK consistency as a mandatory empirical requirement, we utilize it as a dynamic diagnostic tool. In real-world financial time series, the first-order Markov assumption (where $X_{t+1}$ depends only on $X_t$ and $F_t$) is inherently an approximation. Our time-resolved CK analysis (\Cref{sec:results_ck}) reveals that this assumption does not fail uniformly; rather, it breaks down during specific, identifiable market regimes. Spikes in CK divergence serve as valuable structural signals indicating periods where the market exhibits extended memory or where unobserved latent factors dominate the transition dynamics. This type of localized, logical debugging is entirely absent in generic sequence models.

\subsection{Limitations and the Path Forward}
While the proposed framework successfully recovers interpretable time-varying dynamics, its absolute predictive performance remains constrained by the fundamental limits of the empirical setting. 
\begin{itemize}
    \item \textbf{Data Quality and Scale:} The current study is deliberately restricted to a single equity to clearly illustrate the operator diagnostics. Furthermore, it relies heavily on fundamental features that are often artificially interpolated or carried forward over quarterly intervals, naturally diluting the predictive signal-to-noise ratio. 
    \item \textbf{Cross-Asset Dynamics:} The most immediate extension is to scale the neural parameterization to ingest high-frequency, synchronous data across a multi-asset universe. Transitioning from a single-asset state space to a portfolio-level operator is expected to materially improve predictive accuracy by leveraging cross-sectional correlation dynamics, while directly informing downstream applications such as dynamic asset allocation and systematic risk management.
    \item \textbf{Continuous-Time Formulations:} A longer-term direction is to extend the transparent operator view into continuous time (e.g., via neural stochastic differential equations), allowing for irregular observation intervals and a more natural treatment of market microstructure.
\end{itemize}

Ultimately, by restricting a highly flexible neural network to output a mathematically coherent structural form, this framework provides a transparent, calibrated microscope for studying non-stationary market dynamics. It demonstrates that the analytical rigor of classical stochastic processes and the representational power of modern deep learning can be successfully unified.

\section{Conclusion}\label{sec:conclusion}

We introduced a framework for learning time-varying Markov transition
operators in settings where classical empirical estimation fails. In
high-resolution financial discretizations, direct count-based
estimators collapse under sparsity: at $N = 55$ bins, $99.9\%$ of
transition cells contain fewer than five observations, leaving no
viable path to structured operator analysis through tabulation alone.
Our response is to replace empirical counting with neural
parameterization, using a shared-parameter network to estimate
conditional transition rows while preserving the output as an
explicit stochastic operator.

The empirical results support this design. The state-conditioned
model produces operators with mean row heterogeneity
$\bar{\rho} = 0.0073$ (peak $0.014$) across the full time series,
while the state-free ablation collapses to exactly $\rho = 0.0000$
at every timestep---confirming that conditioning on the discretized
Markov state is necessary to obtain genuinely state-dependent
dynamics. Aligning the learned operators with a realized variance
proxy reveals that high-volatility regimes homogenize transition
rows rather than diversify them, with a full-series Pearson
correlation of $r = -0.62$ ($p \approx 10^{-251}$) between row
entropy and $\mathrm{RV}_t$. At $h = 5$, where the directly trained
operators are genuinely concentrated, the Chapman--Kolmogorov
diagnostic identifies specific temporal windows where the
first-order Markov approximation is most strained, with
state-conditioned CK discrepancy running $6\times$ higher than the
state-free baseline and correlating with operator concentration at
$r = -0.533$. These findings are only accessible because the
framework exposes an explicit operator; they are unavailable to any
black-box sequence model.

Predictive performance at the single-asset level is modest and
honestly reported: $\Delta\mathrm{NLL} = +0.025$ at $h = 10$ and
near or below marginal at shorter horizons, consistent with
weak-form efficiency and the limitations of a low-frequency
single-equity feature set. An ongoing extension to a multi-asset
universe with higher-frequency data and substantially larger sample
sizes is expected to improve predictive accuracy materially, while
preserving the operator-level interpretability that is the
contribution of the present work.


\begin{thebibliography}{99}

\bibitem{hamilton1989}
Hamilton, J. D. (1989).
\newblock A new approach to the economic analysis of nonstationary time series and the business cycle.
\newblock \emph{Econometrica}, 57(2), 357--384.

\bibitem{lando2002}
Lando, D., \& Skodeberg, T. M. (2002).
\newblock Analyzing rating transitions and rating drift with continuous observations.
\newblock \emph{Journal of Banking \& Finance}, 26(2), 423--444.

\bibitem{boyd2004}
Boyd, S., \& Vandenberghe, L. (2004).
\newblock \emph{Convex Optimization}.
\newblock Cambridge University Press. 

\bibitem{bishop1994}
Bishop, C. M. (1994).
\newblock Mixture density networks.
\newblock Technical Report NCRG/94/004, Neural Computing Research Group, Aston University.

\bibitem{bengio1995}
Bengio, Y., \& Frasconi, P. (1995).
\newblock An input output HMM architecture.
\newblock \emph{Advances in Neural Information Processing Systems}, 7, 427--434.

\bibitem{salinas2020}
Salinas, D., Flunkert, V., Gasthaus, J., \& Januschowski, T. (2020).
\newblock DeepAR: Probabilistic forecasting with autoregressive recurrent networks.
\newblock \emph{International Journal of Forecasting}, 36(3), 1181--1191.

\bibitem{awiszus2018}
Awiszus, M., \& Rosenhahn, B. (2018).
\newblock Markov chain neural networks.
\newblock \emph{Proceedings of the IEEE Conference on Computer Vision and Pattern Recognition (CVPR) Workshops}.

\bibitem{mettle2022}
Mettle, F. O., Boateng, L. P., Quaye, E. N. B., Aidoo, E. K., \& Seidu, I. (2022).
\newblock Analysis of exchange rates as time-inhomogeneous Markov chain with finite states.
\newblock \emph{Journal of Probability and Statistics}, 2022.

\bibitem{cont2001}
Cont, R. (2001).
\newblock Empirical properties of asset returns: stylized facts and statistical issues.
\newblock \emph{Quantitative Finance}, 1(2), 223--236.


\bibitem{bellemare2017}
Bellemare, M. G., Dabney, W., \& Munos, R. (2017).
\newblock A distributional perspective on reinforcement learning.
\newblock \emph{International Conference on Machine Learning (ICML)}, 449--458.


\bibitem{lim2021}
Lim, B., \& Zohren, S. (2021).
\newblock Time-series forecasting with deep learning: a survey.
\newblock \emph{Philosophical Transactions of the Royal Society A}, 379(2194).

\end{thebibliography}
\end{document}